\documentclass{article}
\usepackage{amsmath, amssymb}
\usepackage{amsthm}
\usepackage{mathtools}
\usepackage{booktabs}
\usepackage{multirow}
\usepackage{threeparttable}

\usepackage{caption}
\usepackage{dsfont}
\usepackage{natbib}
\usepackage{fix-cm}
\usepackage[colorlinks=true, citecolor=blue]{hyperref} 
\usepackage[margin=1in]{geometry}
\usepackage{authblk}
\def\argmin{\mathop{\rm argmin}}

\newcommand{\bA}{\mathbf{A}}

\newcommand{\bC}{\mathbf{C}}

\newcommand{\bI}{\mathbf{I}}
\newcommand{\bK}{\mathbf{K}}

\newcommand{\bU}{\mathbf{U}}
\newcommand{\bV}{\mathbf{V}}

\newcommand{\bX}{\mathbf{X}}

\newcommand{\bk}{\mathbf{k}}

\newcommand{\bv}{\mathbf{v}}

\newcommand{\bx}{\mathbf{x}}

\newcommand{\bbeta}{\boldsymbol{\beta}}
\newcommand{\btheta}{\boldsymbol{\theta}}

\newcommand{\sumps}{\sum_{i\in I_+}}
\newcommand{\summs}{\sum_{j\in I_-}}

\newcommand{\sumk}{\sum_{k=1}^{n}}

\newcommand{\rkhs}{\mathcal{H}_K}

\newtheorem{theorem}{Theorem}           

\newtheorem{proposition}[theorem]{Proposition}

\newtheorem{definition}[theorem]{Definition}

\providecommand{\keywords}[1]{\textit{Keywords:} #1}
\begin{document}

\title{Scaling Up ROC-Optimizing Support Vector Machines}

\author[1]{Gimun Bae}
\author[1]{Seung Jun Shin}

\affil[1]{Department of Statistics, Korea University, Seoul, 02841, Republic of Korea}

\date{}

\maketitle
\begin{abstract}
The ROC–SVM, originally proposed by \citet{Rakotomamonjy2004Optimizing}, directly maximizes the area under the ROC curve (AUC) and has become an attractive alternative of the conventional binary classification under the presence of class imbalance. 
However, its practical use is limited by high computational cost, as training involves evaluating all $O(n^2)$ pairwise terms. To overcome this limitation, we develop a scalable variant of the ROC–SVM that leverages incomplete U-statistics, thereby substantially reducing computational complexity. We further extend the framework to nonlinear classification through a low-rank kernel approximation, enabling efficient training in reproducing kernel Hilbert spaces. Theoretical analysis establishes an error bound that justifies the proposed approximation, and empirical results on both synthetic and real datasets demonstrate that the proposed method achieves comparable AUC performance to the original ROC–SVM with drastically reduced training time.
\end{abstract}


\maketitle

\section{Introduction}
Binary classification is a fundamental problem in machine learning. Given a pair $(\bX, Y)$, where $\bX$ is a p-dimensional predictor and $Y$ is a binary response taking values in $\{-1,1\}$, the goal is to learn a decision function $f$ of $\bX$ that predicts $Y$ by $\hat Y = \mbox{sign}\{f(\bX) \}$. A canonical approach is to choose $f$ that minimizes the classification error, or equivalently, maximizes the accuracy. For instance, the support vector machine (SVM; \citealp{Vapnik1999-VAPTNO}) determines the decision function by maximizing the geometric margin, which effectively aligns with maximizing accuracy \citep{lin2002support}.

However, in imbalanced settings where one class is substantially underrepresented, accuracy can be a misleading measure of performance. Even a trivial classifier that always predicts the majority class can achieve high accuracy while completely failing to detect samples from the minor class. As an alternative, the receiver operating characteristic (ROC) curve is widely used to evaluate classifier performance under class imbalance. By definition, the ROC curve plots the true positive rate (TPR) against the false positive rate (FPR) to summarize classification performance, and the area under the ROC curve (AUC) serves as a popular scalar summary. A classifier with a larger AUC value is generally regarded as having better classification performance.

By defintion, the population AUC of a classifier $f$ is given by
\begin{align} \label{eq:pop.auc}
\mathrm{AUC}(f) = \mathbb{P}\bigl( f(\mathbf{X}^+) > f(\mathbf{X}^-) \bigr),
\end{align}
where $\mathbf{X}^+$ and $\mathbf{X}^-$ denote predictor vectors from the positive and negative classes, respectively.
Unlike accuracy, AUC is less sensitive to class proportions. This robustness has also been examined from a theoretical perspective. \citet{Cortes2003auc} was the first to theoretically investigate the relationship between classification error and AUC, showing that minimizing the error rate does not necessarily lead to the best AUC performance. Furthermore, \citet{rosset2004model} demonstrated that AUC can serve as a more reliable criterion than empirical error, even when the objective is to minimize the misclassification rate.

This motivates us to directly maximize the AUC rather than accuracy. Suppose we are given data $(\bx_i, y_i)\in \mathbb{R} \times \{-1, 1\}$ for $i = 1, \cdots , n$. Let $I_+ = \{i: y_i = 1\}$ and $I_- = \{i: y_i = -1\}$ denote the index sets corresponding to the positive and negative classes, respectively, and define $n_+ = |I_+|$ and $n_- = |I_-|$ so that $n_+ + n_- = n$. Then, the empirical counterpart of \eqref{eq:pop.auc} is given by
\begin{align} \label{eq:sample.auc}
\frac{1}{n_+ n_-} \sum_{i \in I_+} \sum_{j \in I_-} \mathds{1}\{ f(\bx_i) \leq f(\bx_j) \},
\end{align}
where $\mathds{1}\{\cdot\}$ denotes the indicator function. However, the direct optimization of \eqref{eq:sample.auc} is computationally intractable due to the iregularity of the zero–one loss.
To overcome this, the zero–one loss can be replaced with a convex surrogate \citep{Rakotomamonjy2004Optimizing, clemenccon2013empirical, natole2018stochastic, yang2020stochastic}.
Among various choices, the hinge loss is a nautral choice due the popularity of the SVM.
The ROC–SVM \citep{Rakotomamonjy2004Optimizing} solves
\begin{align} \label{eq:ROCSVM}
\min_{f \in \rkhs} \frac{1}{n_+ n_-}\sumps \summs [1 - \{f(\mathbf{x}_{i}) - f(\mathbf{x}_{j})\}]_{+} + \frac{\lambda}{2} \|f\|_{\rkhs}^2
\end{align}
where \( [z]_+ = \max\{z, 0\} \), $\rkhs$ denotes the reproducing kernel Hilbert space (RKHS; \citealp{scholkopf2002learning}) generated by a positive definite kernel $K(\bx, \bx^\prime)$, and $\lambda >0$ is a regularization parameter controlling the complexity of $f$. The ROC–SVM closely resembles the standard SVM, which has led to a variety of extensions \citep[][among many others]{arzhaeva2006linear, tian2011auc, 10.5555/3104482.3104512, ying2016stochastic}.

Despite its conceptual simplicity, computing the ROC–SVM in \eqref{eq:ROCSVM} remains prohibitively intensive for large $n$, as it requires evaluating \( n_+ \times n_- \) pairwise terms, which scales as $O(n^2)$. To address this computational bottleneck, several studies have proposed improvements to enhance the efficiency. For example, \citet{kim2020regularization} developed a  regularization path algorithm that substantially accelerates parameter tuning by exploiting the piecewise linearity of the ROC–SVM solution. However, this approach does not reduce the inherent computational complexity of the ROC–SVM itself.

In this article, we propose an efficient algorithm for solving \eqref{eq:ROCSVM}, even when $n$ is substantially large. We first note that the empirical AUC in \eqref{eq:sample.auc} is a U-statistic, and we leverage the incomplete U-statistic approach of \citet{clemenccon2016scaling}, which significantly reduces the computational burden while keeping the approximation error within a controlled level. While the linear ROC–SVM naturally benefits from this incomplete U-statistic approximation, its kernel version is still computationally demanding due to the size of the kernel matrix which scales as $O(n^2)$. To address this, we further employ a low-rank approximation of the kernel matrix, which not only transforms the kernel ROC–SVM into a linear ROC–SVM in the induced feature space but also substantially reduces the memory requirement.

The remainder of this paper is organized as follows.
Section~\ref{sec:Incomplete U-statistics} introduces incomplete U-statistics and their application to the linear ROC–SVM.
Section~\ref{sec:Scalling Up Kernel ROC-SVM} extends the proposed idea to the kernel ROC–SVM via a low-rank approximation of the kernel matrix and provides its approximation-error bound. Section~\ref{sec:Simulation} presents application to both synthetic and real data sets to illustrate the effectiveness of the proposed method, and Section~\ref{sec:Conclusion} concludes the paper. Technical details are provided in \nameref{sec:Appendix}.

\section{ROC-SVM as an Incomplete U-statistic} \label{sec:Incomplete U-statistics}

\subsection{Incomplete U-statistic}
Let $U_n$ denote the loss part of ROC-SVM in \eqref{eq:ROCSVM}:
\begin{align} \label{eq:erm_roc}
U_n = \frac{1}{n_+ n_-}\sumps \summs [1 - \{f(\mathbf{x}_{i}) - f(\mathbf{x}_{j})\}]_{+}
\end{align}
We note that $U_n$ takes the form of the generalized U-statistic defined as follows.  
\begin{definition} {\bf (Generalized U-statistic)} Suppose we have $J$ independent population, and we let \( \bx_{i,j}\) denote the \(i\)-th i.i.d observations from the \( j \)-th  population where $i = 1, 2, \cdots, n_j$ and $j = 1, 2, \cdots, J$. 
Let $\Lambda_j$ denotes all possible combination of index $(1, 2, \cdots, n_j)$ of a given size $d_j$, where $j = 1, 2, \cdots, J$. 
Then, the generalized U-statistic with a kernel functinon $H: \mathbb{R}^p \times \cdots \times \mathbb{R}^p \mapsto \mathbb{R} $, is defined as
\begin{align} \label{eq:Gen_U_stat}
    U_n = \frac{1}{\prod_{j=1}^{J} \binom{n_j}{d_j}} 
    \sum_{I_1 \in \Lambda_1} \cdots \sum_{I_J \in \Lambda_J} H \left( \bX_{I_1, 1}, \ldots, \bX_{I_J, J} \right).
\end{align}
where $\bX_{I_j, j} = \{\bx_{i,j} \mid i \in I_j\}$ for a given index set $I_j = \{i_{j1}, \cdots, i_{j d_j} \} \subset \{1, 2, \cdots, n_j \}$ for the \(j\)-th sample, $j = 1, 2, \cdots, J$, and $n = \sum_{j=1}^J n_j$.
\end{definition}
The generalized U-statistic \eqref{eq:Gen_U_stat} reduces to the conventional U-statistic when $J = 1$. A well-known example of a generalized U-statistic is the Mann–Whitney statistic \citep{lee2019u} for two-sample location problem. Namely, the Mann–Whitney statistic is a form (upto constant multiplication) of \eqref{eq:Gen_U_stat} with $J = 2$ and $d_j = 1$ with a zero-one kernel $H(x_{i,1}, x_{i,2}) = \mathds{1} \{x_{i,1} < x_{i,2} \}$. In fact, the AUC is equivalent to the Mann–Whitney statistic \citep{yan2003optimizing}, and $U_n$ in \eqref{eq:erm_roc} is a convex relaxation of the sample $1 -$ AUC that replaces the zero-one loss with the hinge loss function. 

The generalized U-statistic involves sum of $\prod_{j=1}^{J} \binom{n_j}{d_j}$ terms, which becomes quickly intractable as \( n_j \) gets large, even with $d_j = 1, j = 1, 2, \cdots, J$. Notably, due to dependence among terms, a small subset can often approximate the full sum without significantly increasing the variance \citep{lee2019u}. This motivates the use of incomplete U-statistics \citep{blom1976some,clemenccon2016scaling} defined in the following, and this greately reduces the computational cost of the U-statistics while preserving a desired level of accuracy.

\begin{definition}{\bf (Incomplete Generalized U-statistic)} 
Let \( \Lambda \) denote the set of all index tuples $(I_1, \ldots, I_J), \forall I_j \in \Lambda_j, j = 1, 2, \cdots J$. 
Given a randomly chosen subset \( \mathcal{D}_B \subseteq \Lambda \) of size B, the incomplete generalized U-statistic~\eqref{eq:Gen_U_stat} is defined as
\begin{align} \label{eq:Incom_gen_U_stat}
    \tilde{U}_{B} = \frac{1}{B} \sum_{(I_1, \ldots, I_J) \in \mathcal{D}_B} 
    H \left( \bX_{I_1, 1}, \ldots, \bX_{I_J, J} \right).
\end{align}
\end{definition} 
The cardinalty of $\mathcal{D}_B$, $B$ is much smaller than that of $\Lambda$. The incomplete U-statistic subtantially reduces the computational cost while preserving the learning rate. Specifically, the incomplete U-statistic $\tilde U_B$ achieves the same convergence rate bound as $U_n$, \( O_{\mathbb{P}}\left(\sqrt{\frac{\log(n)}{n}}\right) \), if $\mathcal{D}_B$ is randomly chosen index tuples from \( \Lambda \) with replacement and $B = O(n)$ \citep{clemenccon2016scaling}.

\subsection{Scaling-up Linear ROC-SVM via Incomplete U-statistic}
Suppose the classification function is linear, that is, $f(\bx) = \alpha + \bbeta^T \bx$. Then, the incomplete ROC–SVM solves
\begin{align} \label{eq:erm_roc_in}
\min_{\bbeta} \frac{1}{B}\sum_{(i,j) \in \mathcal{D}_B} [1 - \bbeta^T (\bx_{i} - \bx_{j})]_{+} + \frac{\lambda}{2} \bbeta^T \bbeta,
\end{align}
where $\mathcal{D}_B$ is a randomly selected subset (with replacement) of $\Lambda = \{(i, j) \mid \forall i \in I_+, j \in I_-\}$ with replacement. We remark that the intercept \( \alpha \) is not identifiable, as it cancels out in the difference  \( f(\mathbf{x}_i) - f(\mathbf{x}_j) \). Nevertheless, it is still required for prediction, and one can determine $\alpha$ to achieve a desired sensitivity or specificity \citep{kim2020regularization}.

Regarding the choice of \( B \), there exists a trade-off between computational efficiency and predictive performance.
 A smaller \( B \) reduces computational cost but increases variance, whereas a larger \( B \) yields more accurate estimates at the expense of higher computational burden. \citet{clemenccon2016scaling} assume \( B \) increases at the rate \( O(n) \) to obtain an approximation error bound of \( O_{\mathbb{P}}\left(\sqrt{\frac{\log(n)}{n}}\right) \). Accordingly, we set \( B = n \), the number of training samples, in subsequent applications in Section \ref{sec:Simulation}.

Finally, we employ the gradient descent algorithm (GDA) to solve \eqref{eq:erm_roc_in}.
Let $\mathcal{L}(\bbeta)$ denote the objective function in \eqref{eq:erm_roc_in}.
The GDA update for $\bbeta$ is given by
\begin{align} \label{eq:GD_update_formula}
\bbeta^{(t+1)}=\bbeta^{(t)}-\eta_{t} \nabla \mathcal{L}(\bbeta^{(t)}),
\end{align}
where
$$
\nabla \mathcal{L} (\bbeta) = \frac{\partial \mathcal{L} (\bbeta)}{\partial (\bbeta)}
= -\frac{1}{B} \sum_{(i,j)\in \mathcal{D}_B} (\bx_i - \bx_j) \cdot \mathds{1}\left\{1 - \bbeta^T (\bx_i - \bx_j) > 0\right\} + \lambda \bbeta.
$$
The step size $\eta_t$ in \eqref{eq:GD_update_formula} can be chosen in several ways.
A simple approach is to fix $\eta_t$ as a small constant, such as $10^{-2}$ or $10^{-3}$. For more efficient optimization, adaptive methods such as the ADAM algorithm have been proposed in the context of stochastic gradient descent \citep{article}.
Among these, we adopt ADAMAX, a variant of the ADAM algorithm based on the infinity norm.
Although \eqref{eq:GD_update_formula} describes a GDA algorithm rather than stochastic gradient descent, the ADAMAX update remains computationally efficient and yields stable solutions. We refer readers to Section 7 of \citet{article} for details on the ADAMAX update.

\section{Scaling-up Kernel ROC-SVM} \label{sec:Scalling Up Kernel ROC-SVM}

\subsection{Low-Rank Linearization of the Kernel ROC-SVM}
By the Representer Theorem \citep{kimeldorf1971some}, the solution to \eqref{eq:ROCSVM} admits the following finite representation:
\begin{align} \label{eq:finite}
f(\bx) = \alpha + \sumk \theta_k K(\bx, \bx_k)
\end{align}
Plugging \eqref{eq:finite} into \eqref{eq:ROCSVM}, the kernel ROC-SVM solves
\begin{align}\label{eq:kernel-ROCSVM}
    \min_{\btheta}  \frac{1}{n_+ n_-} \sum_{i\in I_+}\sum_{j\in I_-}[1-\btheta^T( \bk_{i} - \bk_j)]_{+} + \frac{\lambda}{2} \btheta^\top\bK\btheta
\end{align}
where $\btheta = (\theta_1, \cdots, \theta_n)^T$, \(\bK\) is the $n$-dimensional kernel matrix whose $(l,m)$th element is \(K (\bx_l,\bx_m)\), and $\bk_i$ is the $i$th column of $\bK$. 

Despite the use of the incomplete U-statistic, solving \eqref{eq:kernel-ROCSVM} remains challenging, particularly when $n$ is excessively large, because the kernel matrix $\bK$ in the penalty term requires $O(n^2)$ memory and computation. To address this issue, we employ the idea of low-rank linearization \citep{zhang2012scaling}.
Specifically, one can transform \eqref{eq:kernel-ROCSVM} into an equivalent but computationally simpler linear ROC–SVM.
\begin{proposition} 
\label{prop:trans}
For a given kernel matrix $\bK$, suppose we have $\bK = \bV \bV^T$, \eqref{eq:kernel-ROCSVM} is equivalent to solve the following linear ROC-SVM
\begin{align*}
    \min_{\bbeta}
    \frac{1}{n_+ n_-} \sum_{i\in I_+}\sum_{j\in I_-} \left[ 1 - \bbeta^T \left(\bv_i - \bv_j \right) \right]_+ 
    + \frac{\lambda}{2} \boldsymbol{\beta}^\top\boldsymbol{\beta}
\end{align*}
where $\bv_i$ denotes the $i$th row of $\bV$. 
\end{proposition}

The proof of Proposition~\ref{prop:trans} is provided in the Appendix. A straightforward solution of $\bV$ can be obtained from the eigen-decomposition of $\bK$. Namely, we have 
$$
\mathbf{V}=\bU \boldsymbol{\Lambda}^{1/2}.
$$
where \(\boldsymbol{\Lambda}\in \mathbb{R}^{n \times n}\) is a diagonal matrix whose diagonal elements are eigenvalues of \(\bK\) arranged in descending order and \(\mathbf{U} \in \mathbb{R}^{n \times n}\) is the orthogonal matrix whose columns are the corresponding eigenvectors of \(\bK\). 
However, this approach still requires computing the full kernel matrix $\bK$, and the resulting \( \bV \in \mathbb{R}^{n \times n} \) provides no computational advantage. 

To tackle this, we apply the Nyström approximation \citep{williams2000using} to obtain \( \tilde \bV \in \mathbb{R}^{n \times d} \), a low-rank approximation of $\bV$. Nyström Approximation begins by selecting $d (\ll n)$ landmark samples from $\bx_i, i = 1, 2, \cdots, n$, and then compute the corresponding kernel matrix denoted by $\bK_{\text{selected}} \in \mathbb{R}^{d\times d}$. The Landmark samples can be chosen through different strategies: uniform random sampling is the simplest approach, whereas sophisticated methods such as K-means clustering have been proposed to reduce approximation error \citep{zhang2010clustered}. For the theoretical analysis in Section~\ref{subsec:Approximation Error Bound}, we assume the uniform random sampling for simplicity. In our simulation studies, however, we employ the stratified random sampling to account for class imbalance, and in real-data applications we adopt the K-means–based selection strategy, as our results are not overly sensitive to the specific choice of landmark selection method.

Given $d$ landmark samples, we construct 
\begin{align} \label{eq:Nystrom_V}
    \tilde \bV = \bC(\bK_{\text{selected}}^{\dagger})^{1/2},
\end{align}
where $\bA^{\dagger}$ denotes Moore-Penrose inverse of $\bA$ and $\bC\in \mathbb{R}^{n\times d}$ is the kernel matrix computed between all training samples $\bx_i, i = 1, 2, \cdots, n$ and the $d$ landmark samples. Compared with the exact formulation, this approximation in \eqref{eq:Nystrom_V} only requires computing \( d \times d \) kernel matrix \( \bK_{\text{selected}} \) (and its inverse) and an \( n \times d \) matrix \( \mathbf{C} \). Unless \( d \) is chosen exesslive large, this procedure is substantially more efficient. 

The kernel ROC-SVM in \eqref{eq:kernel-ROCSVM} can therefore be approximated by the following $d$-dimensional linear ROC-SVM: 
\begin{align}  \label{eq:kernel-ROCSVM-incomp-lin}
    \min_{\boldsymbol{\beta}} \frac{1}{B}\sum_{(i,j)\in \mathcal{D}_B}[1- \bbeta^T (\tilde{\bv}_i - \tilde{\bv}_j)]_{+} + \frac{\lambda}{2} \boldsymbol{\beta}^\top\boldsymbol{\beta},
\end{align}
where $\tilde{\bv}_i$ denotes the $i$th row of $\tilde \bV$.

For the prediction, we first compute the test feature matrix $\tilde{\bV}^{\text{ts}} = \bC^{\text{ts}}(\bK_{\text{selected}}^{\dagger})^{1/2}$, where \(\bC^{\text{ts}} \in \mathbb{R}^{n^{\text{ts}}\times d}\) is defined analogously to $\bC$ using the test samples \(\bx_i^{\text{ts}, i=1,2,\ldots,n^{\text{ts}}}\). Then, the \(i\)th row \(\tilde{\bv}_i^{\text{ts}}\) of \(\tilde{\bV}^{\text{ts}}\) serves as the transformed feature vector for the \(i\)th test observation, whose predicted value is $\hat y_{i}^{\text{ts}} = \hat \alpha + \hat \bbeta^T \bv_{i}^{\text{ts}}$, where $\hat \alpha$ and $\hat \bbeta$ are the parameter estimates from the training data.

\subsection{Approximation Error Bound}
\label{subsec:Approximation Error Bound}
As seen in \eqref{eq:erm_roc}, the ROC–SVM loss can be expressed as a U-statistic with the following kernel:
\begin{align}
    \label{def:H}
    H_f(\bx_i, \bx_j) = [1 - \{f(\bx_i) - f(\bx_j)\}]_+.
\end{align}
We approximate this loss by its incomplete counterpart and construct $f(\bx)$ using the Nyström approximation, which inevitably introduces an approximation error. In what follows, we show that the difference between the empirical losses—with and without the approximation—can be bounded with high probability.

We approximate the loss with the incomplete counterpart and construct $f(\bx)$ using the Nyström approximation, which results approximation error. In what follows, we show that the difference in empirical loss—with and without the approximation—can be bounded with high probability.

For clarity, we introduce some notation. Let \(U_n(f)\) denote the full U-statistic-based loss function~\eqref{eq:erm_roc} with decision function $f$, and \(\tilde U_B(f)\) be the corresponding incomplete U-statistic based on $B$ randomly selected pairs.
A baseline approach is to optimize the full U-statistic \( U_n(f) \) without applying the Nyström approximation.
That is, the method solves \eqref{eq:ROCSVM}, where we write $\mathcal{H} := \mathcal{H}_K$ for simplicity and denote its solution by $\hat{f}$.
In our approach, the model is formulated by solving the linearized problem \eqref{eq:kernel-ROCSVM-incomp-lin}. We remark that Proposition~\ref{prop:trans} remains valid even when the problem is defined using the incomplete U-statistic and the Nyström approximation, since its proof does not depend on either assumption. Thus, the problem \eqref{eq:kernel-ROCSVM-incomp-lin} can be equivalently expressed as
\begin{align}
\label{eq:kernel-ROCSVM-incomp}
    \min_{f \in  \mathcal{\tilde H}}  \frac{1}{B} \sum_{(i,j)\in \mathcal{D}_B}[1-( f(\bx_i)- f(\bx_j))]_{+} + \frac{\lambda}{2} \|f\|_{\mathcal{\tilde H}}^2,
\end{align} 
where $\mathcal{\tilde H}$ is the RKHS induced by the Nyström-approximated kernel matrix $\tilde{\bK}$ defined as $\tilde{\bK} = \tilde{\bV}\tilde{\bV}^{\top}$. 
We note that \(\tilde{\mathcal{H}}\) is well-defined because \(\bK\) is positive definite and \(\tilde{\bK}\) is positive semi-definite. We further denote its solution by $\tilde{f}_B$, which, by the Representer Theorem, 
has the same form as in~\eqref{eq:finite}, with $K(\bx_i,\bx_k)$ replaced by $\tilde K(\bx_i,\bx_k)$, the $(i,k)$-th entry of $\tilde{\bK}$.

Let \(\lambda_{\min}\) denote the minimum eigenvalue of \(\bK\), and let \(\tilde \lambda^{+}_{\min}\) denote the smallest nonzero eigenvalue of \(\tilde \bK\). Define the kernel-approximation error by $\mathcal{E} := \bK - \tilde{\bK}$, and 
let $\mathcal{E}_2 := \|\mathcal{E}\|_2$ denote its spectral norm. We further define the random variable $N_{\max}$ as the maximum number of times a sample is selected in $\mathcal{D}_B$. For the RKHS norm, let $K_{\max}$ denote the supremum of $\|K(\cdot, \bx_i)\|_{\mathcal{H}}$ over all observed data:
\[
K_{\max} = \sup_{1 \le i \le n} \|K(\cdot, \bx_i)\|_{\mathcal{H}},
\] 
where the kernel $K(\bx, \bx^\prime)$ is assumed to be bounded. Let $C$ and $\tilde{C}_B$ be upper bounds on the RKHS norms of the minimizers $\hat{f}$ and $\tilde{f}_B$, respectively; that is, 
$\|\hat{f}\|^2_{\mathcal{H}} \le C$ and $\|\tilde{f}_B\|^2_{\tilde{\mathcal{H}}}\le \tilde{C}_B$. We also denote $C_{\max}$ a finite constant related to $C$, whose precise definition will be specified in the proof. 

We now establish the following theorem, which states that the approximation error is bounded with high probability.
The proof is given in the Appendix.
\begin{theorem} \label{thm:bound}
    Given the observed data, for all $\delta \in (0,1)$, with probability at least 1-$\delta$,
\begin{align*}
    |U_n(\hat{f})-\tilde U_B(\tilde{f}_B)|\le A_1\sqrt{\frac{\log(4/\delta)}{B}} + \frac{N_{\max}}{B}\left(A_2\mathcal{E}_2 \sqrt{n}+A_3\sqrt{\frac{N_{\max}\mathcal{E}_2n^{3/2}}{B}}\right),
\end{align*}
where 
\[
A_1 = \frac{1+2K_{\max}\sqrt{C_{\max}}}{\sqrt{2}}, \quad 
A_2 = \sqrt{\frac{\tilde{C}_B}{\tilde{\lambda}_{\min}^+}}, \quad
A_3 = K_{\max}\sqrt{\tfrac{2}{\lambda}}
\left(\sqrt{\tfrac{ C }{ \lambda_{\min} } }+\sqrt{\tfrac{\tilde{C}_B}{\tilde{\lambda}_{\min}^+}}\right)^{1/2}.
\]
\end{theorem}
We make several remarks on Theorem~\ref{thm:bound}. The first term corresponds to the error introduced by the incomplete U-statistic and converges to zero at the rate $O(1/\sqrt{B}) = O(1/\sqrt{n})$, since $B=O(n)$. 
The second term arises from the Nyström approximation.
By Bernstein’s inequality, one can show that $N_{\max} = O_{\mathbb{P}}(\log n)$ when 
$\min(n_+, n_-) = O(n)$; that is, when no class dominates the other.
Furthermore, \citet{drineas2005nystrom} showed that $\mathcal{E}_2 = O(n/\sqrt{d})$, where $d$ is the target rank. 
Hence, if $d=o(n)$, then $\mathcal{E}_2 = o(n)$, and consequently the second term is of order $o(\sqrt{n} \log n)$. Specifically, if we set $d = n/\log n$, then $\mathcal{E}_2 = O(\sqrt{n\log n})$, and consequently 
the second term is of order $O((\log n)^{3/2})$. Although the upper bound diverges asymptotically, its growth rate $o(\sqrt{n} \log n)$ is subpolynomial, and therefore justifing the use of our approximation in practice.

\section{Application} \label{sec:Simulation}

\subsection{Synthetic Data}
We consider two illustrative scenarios to evaluate the efficiency of the proposed method.
The data are generated from the following binary classification model:
$$
y_i = \mbox{sign}\{\alpha + f(\bx_i) + \epsilon_i\}, ~~ i = 1, 2, \cdots, n,
$$
where the bivariate predictor $\bx_i \sim N_2(\mathbf{0}, \bI)$ and the random error $\epsilon_i \sim N(0, 1)$. The offset  parameter $\alpha$, which controls the class imbalance, is chosen so that approximately 80\% of the responses $y_i$ are negative.

We consider two different forms of the classification function $f$. 
\begin{itemize}
   \item Linear Model -- $f(\bx_i) = \bbeta^T \bx_i$, where $\bbeta - (1, 1)^T$. 
   \item Radial Model -- $f(\bx_i) = \| \bx_i \|^2$
\end{itemize}
We generate training samples of sizes $n_{\text{train}} \in \{5, 50, 100\}\times 10^3$ while fixing the test sample size at $_{\text{test}} = 25 \times \times 10^3$. 

To evaluate the finite sample performance, we compute the empirical AUC as
\[
\hat{\text{AUC}} = \frac{1}{n_+n_-} \sum_{i \in I_+} \sum_{j \in I_-} \mathds{1}(\hat{f}(\bx_i) > \hat{f}(\bx_j)),
\]
where $\hat{f}(\bx)$ denotes the estimated classification function. As a reference, we also compute the population AUC via Monte Carlo approximation, which is feasible because the true function $f$ is known in the simulation setting.

For the linear model, we train the ROC–SVM using both the full U-statistic risk and the incomplete U-statistic risk, hereafter referred to as the full and incomplete models, respectively. The regularization parameter is selected via cross-validation. We then compare their performance to evaluate the effect of the approximation. 

\begin{table*}[ht]%
\centering
\caption{Comparison of averaged training times and sample AUCs (over 50 independent repetitions) 
between the full and incomplete models when the true classification function is linear.
\label{tab:sim_linear}}
\begin{tabular*}{\textwidth}{@{\extracolsep\fill}rcc c cc@{\extracolsep\fill}}
      \toprule
                         & \multicolumn{2}{c}{Computing Time (in sec.)}  && \multicolumn{2}{c}{AUC(\%)}\\
      \multicolumn{1}{c}{$n_{\text{train}}$} 
                        & \multicolumn{1}{c}{\textbf{Full}} & \multicolumn{1}{c}{\textbf{Incomplete}} 
                       && \multicolumn{1}{c}{\textbf{Full}} & \multicolumn{1}{c}{\textbf{Incomplete}}  \\
      \midrule
      5{,}000 & 196.10  (7.75) & 2.69 (0.15) && 90.370 (0.88) & 90.370 (1.14) \\
      10{,}000 & 819.10 (31.41) & 3.02 (0.21) && 90.374 (0.37) & 90.373 (0.54)\\
      50{,}000 & \multicolumn{1}{c}{--} & 8.03 (0.90) && \multicolumn{1}{c}{--} & 90.375 (0.22)\\
      100{,}000 & \multicolumn{1}{c}{--} & 10.26 (0.24) && \multicolumn{1}{c}{--} & 90.376 (0.10)\\
      \bottomrule
    \end{tabular*}
    \begin{tablenotes}
\item[$^\dagger$] The population AUC ($\mathrm{AUC}_{\text{true}}$) is 90.377\%. 
The symbol “--” indicates that model fitting was computationally infeasible for the given sample size. 
Numbers in parentheses denote standard errors.
\end{tablenotes}
\end{table*}
Table~\ref{tab:sim_linear} reports the average computing time and sample AUC over 50 independent replications for the linear model, whose population AUC is 0.90377. The results show that the incomplete model substantially reduces computation time compared with the full model, supporting the validity of the incomplete approximation. This advantage becomes more evident as $n$ increases: for instance, the full model fails to run when the training size exceeds 10{,}000 due to memory limitations, whereas the incomplete model scales to about 100{,}000 training samples within several seconds. Nevertheless, the incomplete model exhibits virtually no loss in sample AUC, clearly demonstrating the practical effectiveness of the proposed method.

For the nonlinear model, we train the ROC–SVM with a radial kernel. In this case, the full model cannot be computed because of its prohibitive cost for large $n$ (e.g., $n = 500,000$). Therefore, we fit only the incomplete model and compare it with its Nyström-approximated counterpart ($d=300$). This comparison allows us to assess the efficiency of the Nyström approximation. Table~\ref{tab:sim_nonlinear} summarizes the results.

\begin{table*}[ht]%
\centering
   \caption{Comparison of averaged training times and sample AUCs (over 50 independent repetitions) 
between the incomplete models with and without the Nyström approximation for the nonlinear model. \label{tab:sim_nonlinear}}
\begin{tabular*}{\textwidth}{@{\extracolsep\fill}rcc c cc@{\extracolsep\fill}}
      \toprule
                         & \multicolumn{2}{c}{Computing Time (in sec.)}  && \multicolumn{2}{c}{AUC(\%)}\\
      \multicolumn{1}{c}{$n_{\text{train}}$} 
                        & \multicolumn{1}{c}{\textbf{Without Nyström}} & \multicolumn{1}{c}{\textbf{Nyström}} 
                       && \multicolumn{1}{c}{\textbf{Without Nyström}} & \multicolumn{1}{c}{\textbf{Nyström}}\\
      \midrule      
      5{,}000   & 10.21 (0.30) & 4.64 (0.25) && 90.947 (0.53) & 90.921 (0.69)\\
      10{,}000  & 19.81 (2.84) & 4.60 (0.52) && 90.863 (2.27) & 90.927 (0.64)\\
      50{,}000  & \multicolumn{1}{c}{--} & 5.19 (0.14) && \multicolumn{1}{c}{--} & 90.936 (0.61)\\
      100{,}000 & \multicolumn{1}{c}{--} & 5.87 (0.14) && \multicolumn{1}{c}{--} & 90.926 (0.64)\\
      \bottomrule
    \end{tabular*}
    \begin{tablenotes}
\item[$^\dagger$] The population AUC ($\mathrm{AUC}_{\text{true}}$) is 90.985\%. 
The symbol “--” indicates that model fitting was computationally infeasible for the given sample size. 
Numbers in parentheses denote standard errors.\end{tablenotes}
\end{table*}

We observe that the incomplete U-statistic method alone cannot handle training sets larger than 10{,}000 samples, whereas the Nyström approximation drastically reduces computational burden and enables model fitting with training sizes up to about 100{,}000 within a few seconds. In terms of AUC, the Nyström approximation introduces only a minor loss, slightly increasing the gap from the population AUC (0.90985); nevertheless, the empirical AUC remains close to the population value, indicating that predictive performance is well preserved.

\subsection{Real Data}
We further evaluate the performance of our methods on three datasets for imbalanced classification.
The first dataset is the Skin Segmentation dataset \citep{skin_segmentation_229}, which contains 245,057 observations with three predictors and a positive-class proportion of 20.8\%.
Because the dataset is large and the class ratio ($n_+/n$) is not extremely small, fitting the model on the full data is infeasible on a personal computer due to memory constraints.
Therefore, we randomly sample 100,000 observations while preserving the original class distribution.
The second dataset is the Bank Marketing dataset \citep{Moro2014ADA}, which includes 41,188 observations with 19 predictors and a positive-class proportion of 11.3\%. The third dataset is the Accelerometer Gyro Mobile Phone dataset \citep{accelerometer_gyro_mobile_phone_755}, containing 31,991 observations with six predictors and a positive-class proportion of 1.8\%.

For the real-data experiments, each dataset is randomly split into training and test sets.
We then apply the proposed linear and kernel ROC–SVM models using the incomplete U-statistic approximation.
For the kernel model, we further employ the Nyström approximation with $d=300$. The entire procedure is repeated 30 times, and we record the training time and empirical AUC on the test sets. The regularization parameter $\lambda$ is selected via cross-validation. Table~\ref{tab:real_data_result} reports the average training time and test AUC obtained at the optimal value of $\lambda$. Both proposed models train within seconds while achieving high test AUC.
In particular, the kernel model consistently attains higher AUC than the linear model, and for the Skin Segmentation and Accelerometer datasets it even trains faster.
Although the kernel model trains somewhat more slowly on the Bank Marketing dataset and exhibits a larger standard error in AUC for the Accelerometer dataset, the overall computational cost remains practical and the classification performance is strong. In summary, the original (non-approximated) ROC–SVM is computationally infeasible—requiring hundreds of seconds even for smaller subsamples—whereas our proposed approach achieves competitive accuracy within a fraction of the time.

\begin{table*}[ht]%
  \centering
\caption{Training time and test AUC on the three real datasets, averaged over 30 random train–test splits. 
  The regularization parameter $\lambda$ is selected via cross-validation.
\label{tab:real_data_result}}
\begin{tabular*}{\textwidth}{@{\extracolsep\fill}lcccccc@{\extracolsep\fill}}
      \toprule
    \textbf{Dataset} & $n$ & $p$ & $n_+/n (\%)$ &\textbf{Kernel} & \textbf{Time} & \textbf{AUC}(\%) \\
    \toprule
    \multirow{2}{*}{\shortstack{Skin Segmentation}} 
        &\multirow{2}{*}{100,000}&\multirow{2}{*}{3}& \multirow{2}{*}{20.8}
            & Linear & 8.38 (2.09) & 94.64 (1.96) \\
        &&& & Radial & 5.82 (0.15) & 98.53 (1.40) \\ \midrule
     \multirow{2}{*}{\shortstack{Bank Marketing}} 
        &\multirow{2}{*}{41,188}&\multirow{2}{*}{18}& \multirow{2}{*}{11.3}
            & Linear  & 0.43 (0.04) & 74.71 (9.25) \\
        &&& & Radial & 46.33 (9.61) & 76.04 (7.94)  \\ \midrule
     \multirow{2}{*}{\shortstack{Accelerometer}} 
        &\multirow{2}{*}{31,991}&\multirow{2}{*}{6}& \multirow{2}{*}{$~$1.8}
            & Linear  & 3.54 (0.12) & 88.53 (9.15) \\
        &&& & Radial & 1.60 (0.38) & 89.92 (21.61)  \\
    \bottomrule
\end{tabular*}
\begin{tablenotes}
\item[$^\dagger$] Numbers in parentheses denote standard errors. AUC values are reported as percentages, with standard deviations shown without the $10^{-3}$ factor.
\end{tablenotes}
\end{table*}

\section{Concluding Remarks}
\label{sec:Conclusion}
In this work, we addressed the challenge of scaling ROC–SVM to large datasets, where training involves evaluating $O(n^2)$ pairwise terms, rendering the standard approach impractical. By leveraging incomplete U-statistics and a low-rank kernel approximation, we proposed a computationally efficient algorithm that reduces the overall complexity to $O(n)$. Beyond algorithmic development, we also provided theoretical justification by deriving an approximation bound for the proposed approach.

Empirical results on simulated and real datasets demonstrate that the proposed method consistently achieves competitive AUC compared with the original ROC–SVM, while substantially reducing computational cost.
These findings confirm the feasibility of efficient ROC-optimizing classification, a task that has traditionally been constrained by computational limitations.

Nevertheless, our approach has room for further improvement.
For instance, extending the theoretical analysis beyond the hinge loss remains challenging, as many surrogate losses fail to satisfy global Lipschitz continuity—a key property underlying our proofs.
In addition, our method is currently a batch algorithm and therefore not well suited for streaming data.
An interesting direction for future work is to extend our framework to an online setting, for example, by incorporating stochastic gradient descent. Such an extension is particularly challenging for the kernel version but warrants further investigation.

\bibliographystyle{unsrtnat}
\bibliography{references}

\section*{Appendix}
\label{sec:Appendix}
The Appendix provides the proofs of Proposition~\ref{prop:trans} and Theorem~\ref{thm:bound} presented in the main article.
\subsection*{A.1 Proof of the proposition~\ref{prop:trans}}
\label{subsec:proof of the proposition}
\begin{proof} 
The loss function $[1-\btheta^\top\{\bk_i - \bk_j\}]_{+}$ in \eqref{eq:kernel-ROCSVM} can be equivalently written as
$[1 -\btheta^\top\bV(\bv_i - \bv_j)]_{+}$,
where $\bv_i$ denotes the $i$th row of $\bV$. 
Similarly, the penalty term $\tfrac{\lambda}{2}\btheta^\top \bK \btheta$ can be expressed as
\[
\frac{\lambda}{2} \btheta^\top \bV \bV^{\top} \btheta
= \frac{\lambda}{2} (\bV^{\top}\btheta)^\top (\bV^{\top}\btheta).
\]
By reparametrizing as $\bbeta = \bV^\top\btheta \in \mathbb{R}^{d \times 1}$, the objective function can be rewritten as
\begin{align*}`
 \frac{1}{n_+n_-} \sumps \summs [1 - \btheta^\top \bV (\bv_i - \bv_j)]_{+} + \frac{\lambda}{2} (\bV^{\top}\btheta)^\top (\bV^{\top}\btheta) = \frac{1}{n_+n_-} \sumps \summs [1 - \bbeta^{\top}(\bv_i - \bv_j)]_{+} + \frac{\lambda}{2} \bbeta^\top \bbeta.
\end{align*}
This coincides with the objective function of a linear ROC-SVM with respect to the transformed features $\bv_i$.
\end{proof}

\subsection*{A.2 Proof of the theorem~\ref{thm:bound}}
\label{subsec:proof of the theorem}
\begin{proof} 
We begin by introducing the parameter vectors 
$\hat{\btheta} = (\hat{\theta}_1, \ldots, \hat{\theta}_n)^{\top}$ 
and $\tilde{\btheta}_B = (\tilde{\theta}_{B,1}, \ldots, \tilde{\theta}_{B,n})^{\top}$ from $\hat{f}$ and $\tilde{f}_B$.
By the Representer Theorem, $\hat{f}$ and $\tilde{f}_B$ can be represented in terms of $\hat{\btheta}$ and $\tilde{\btheta}_B$:
\begin{align*}
    \hat{f}(\bx) &= \sum_{k=1}^n \hat{\theta}_k K(\bx, \bx_k), \quad 
    \tilde{f}_B(\bx) = \sum_{k=1}^n \tilde{\theta}_{B,k} \tilde K(\bx, \bx_k).
\end{align*}
Here, we set the intercept term to zero without loss of generality, since it cancels out in pairwise difference.

We next establish upper bounds on the $\ell_2$ norms of $\hat{\btheta}$ and $\tilde{\btheta}_B$ by first deriving their RKHS norm bounds. 
These RKHS norm bounds follow directly from the definitions of $\hat{f}$ and $\tilde{f}_B$, 
the minimizers of the regularized loss function. 
Note that these optimization problems can be equivalently expressed in constrained form:
\begin{align*}
\hat{f} &= \argmin_{f \in \mathcal{H}} \; \frac{1}{n_+ n_-}\sumps \summs [1 - \{f(\mathbf{x}_{i}) - f(\mathbf{x}_{j})\}]_{+} 
\quad \text{subject to} \quad \| f \|^2_{\mathcal{H}} \leq C , \\
\tilde{f}_B &= \argmin_{f \in \tilde{\mathcal{H}}} \; \frac{1}{B}\sum_{(i,j) \in \mathcal{D}_B} [1 - \{f(\mathbf{x}_{i}) - f(\mathbf{x}_{j})\}]_{+} 
\quad \text{subject to} \quad \| f \|^2_{\tilde{\mathcal{H}}} \leq \tilde C .
\end{align*}
From these constraints, it follows that the minimizers satisfy
\[
\| \hat{f} \|_{\mathcal{H}}^2 \le C,
\qquad
\| \tilde{f}_B \|_{\tilde{\mathcal{H}}}^2 \le \tilde C_B,
\]
where, by the reproducing property,
\[\| \hat{f} \|_{\mathcal{H}}^2 = \| \hat{\btheta} \|_{\mathcal{H}}^2 
= \hat{\btheta}^{\top} \bK \hat{\btheta}, \qquad \| \tilde{f}_B \|_{\tilde{\mathcal{H}}}^2 = \| \tilde{\btheta}_B \|_{\tilde{\mathcal{H}}}^2 
= \tilde{\btheta}_B^{\top} \tilde\bK \tilde{\btheta}_B.\]  Therefore, $\| \hat{\btheta} \|_{\mathcal{H}}^2$ and $\| \tilde{\btheta}_B \|_{\tilde{\mathcal{H}}}^2$ are bounded by $C$ and $\tilde C_B$, respectively. 
From these RKHS norm bounds and the Rayleigh quotient inequality, we have
\[
\| \hat{\btheta} \|_2 \;\leq\; \sqrt{\frac{ \| \hat{\btheta} \|_{\mathcal{H}}^2}{ \lambda_{\min} }} \;\leq\; \sqrt{ \frac{ C }{ \lambda_{\min} } },
\]
since $\bK$ is symmetric positive definite.
Although $\tilde{\bK}$ is only positive semi-definite, the Rayleigh quotient 
can still be applied if we assume that 
$\tilde{\btheta}_B \perp \text{null}(\tilde{\bK})$. 
This assumption is trivial, because any component of 
$\tilde{\btheta}_B$ lying in $\text{null}(\tilde{\bK})$ 
(denote it by $\tilde{\btheta}_{B,\perp}$) does not affect the solution: it vanishes under multiplication with $\tilde{\bK}$ in the decision function $\tilde{f}_B$.
Therefore, the Rayleigh quotient bound on $\mathrm{Range}(\tilde{\bK})$ yields
\[
\|\tilde{\btheta}_B\|_2 
\;\le\; \sqrt{\frac{\|\tilde{\btheta}_B\|_{\tilde{\mathcal{H}}}^2}{\tilde{\lambda}_{\min}^+}}
\;\le\; \sqrt{\frac{\tilde{C}_B}{\tilde{\lambda}_{\min}^+}}.
\]

We now turn to bounding $|U_n(\hat{f})-\tilde U_B(\tilde{f}_B)|$, which is the main goal of this proof.
By the triangle inequality,
\begin{align*}
    |U_n(\hat{f})-\tilde U_B(\tilde{f}_B)|  
    &\leq |U_n(\hat{f})-\tilde U_B(\hat{f}_B)|+|\tilde U_B(\hat{f}_B)-\tilde U_B(\tilde{f}_B)|,
\end{align*}
where $\hat{f}_B$ denotes the incomplete counterpart of $\hat{f}$.  
That is, 
\begin{align}
    \label{eq:kernel_ROC_SVM_incomp}
    \hat{f}_B= \argmin_{f \in \mathcal{H}}  \frac{1}{B} \sum_{(i,j)\in \mathcal{D}_B}[1-( f(\bx_i)- f(\bx_j))]_{+} + \frac{\lambda}{2} \|f\|^2_{\mathcal{H}}.
\end{align}
Similarly to $\hat{f}$, $\|\hat{f}_B\|_{\mathcal{H}}$ is bounded by a constant $\sqrt{C_B}$, and we define $C_{\max} = \max(C, C_B)$.
The corresponding $\hat{\btheta}_B$ can also be obtained from
\begin{align}
    \label{def:hat_f_b}
     \hat{f}_B(\bx) =\sumk \hat{\theta}_{B,k} K(\bx, \bx_k).
\end{align}

For the first term $|U_n(\hat{f})-\tilde U_B(\hat{f}_B)|$, since $\hat{f}$ is the minimizer of $U_n(f)$ over $\mathcal{H}$ and $\hat{f}_B \in \mathcal{H}$, we have $U_n(\hat{f}) \le U_n(\hat{f}_B)$. 
Thus, if $\tilde U_B(\hat{f}_B) < U_n(\hat{f})$, then
\[
|U_n(\hat{f})-\tilde U_B(\hat{f}_B)| 
= U_n(\hat{f})-\tilde U_B(\hat{f}_B) 
\le U_n(\hat{f}_B)-\tilde U_B(\hat{f}_B) 
= |U_n(\hat{f}_B)-\tilde U_B(\hat{f}_B)|.
\]
If instead $U_n(\hat{f}) \leq \tilde U_B(\hat{f}_B)$, then
\[
|U_n(\hat{f})-\tilde U_B(\hat{f}_B)|
= \tilde U_B(\hat{f}_B)-U_n(\hat{f}) 
\le \tilde U_B(\hat{f})-U_n(\hat{f}) 
= |\tilde U_B(\hat{f})-U_n(\hat{f})|,
\]
since $\hat{f}_B$ is the minimizer of $\tilde U_B(f)$ over $\mathcal{H}$ and thus $\tilde U_B(\hat{f}_B) \le \tilde U_B(\hat{f})$.
Therefore,
\begin{align}
\label{proof:first_term}
|U_n(\hat{f})-\tilde U_B(\hat{f}_B)|
\le \sup_{f \in \{\hat{f},\hat{f}_B\}} |\tilde U_B(f)-U_n(f)|.
\end{align}
To derive a probability bound for $\sup_{f \in \{\hat{f},\hat{f}_B\}} |\tilde U_B(f)-U_n(f)|$, we introduce the random sequence, $((\zeta_k(I))_{I \in \Lambda})_{1 \le k \le B}$, where $\Lambda$ is defined as $\{(i,j)| i \in I_+, j \in I_-\}$ and $\zeta_k(I)$ is equal to $1$ if  
$I=(I_1,I_2) \in \Lambda$ has been selected at the $k$-th draw and $0$ otherwise: For $k \in \{1,2,\ldots,B\}$, each $(\zeta_k(I))_{I \in \Lambda}$ are i.i.d. random vectors following a 
$\mathrm{Multinomial}\!\left(1;\tfrac{1}{|\Lambda|},\ldots,\tfrac{1}{|\Lambda|}\right)$ distribution. Then, 
\[    \tilde U_B(f)=\frac{1}{B}\sum_{k=1}^{B}\sum_{I \in \Lambda} \zeta_k(I)H_f(\bx_{I_1},\bx_{I_2}), \quad 
    U_n(f) =\frac{1}{|\Lambda|}\sum_{I \in \Lambda}H_f(\bx_{I_1},\bx_{I_2}),\]
where $H_f(\bx_{I_1},\bx_{I_2})$ denotes the kernel function~\eqref{def:H}.
Given the observed data,
\begin{align*}
P&\left(\sup_{f \in \{\hat{f},\hat{f}_B\}} |\tilde U_B(f)-U_n(f)| > \eta \,\Big|\,(\bx_{I_1},\bx_{I_2})_{(I_1,I_2) \in \Lambda}\right) \\
& \le \sum_{f \in \{\hat{f},\hat{f}_B\}} 
P\left(\left|\frac{1}{B}\sum_{k=1}^{B}\sum_{I \in \Lambda} \zeta_k(I)H_f(\bx_{I_1},\bx_{I_2})
- \frac{1}{|\Lambda|}\sum_{I \in \Lambda}H_f(\bx_{I_1},\bx_{I_2})\right| > \eta \,\Big|\,(\bx_{I_1},\bx_{I_2})_{(I_1,I_2) \in \Lambda}\right),
\end{align*}
where 
\[
E_{\zeta}\left[\frac{1}{B}\sum_{k=1}^{B}\sum_{I \in \Lambda} \zeta_k(I)H_f(\bx_{I_1},\bx_{I_2})\,\Big|\,(\bx_{I_1},\bx_{I_2})_{(I_1,I_2) \in \Lambda}\right]
= \frac{1}{|\Lambda|}\sum_{I \in \Lambda}H_f(\bx_{I_1},\bx_{I_2}).
\]
Note that for each $k$,
\begin{align*}
    \sum_{I \in \Lambda} \zeta_k(I)H_f(\bx_{I_1},\bx_{I_2}) 
    &\le \sup_{f \in \{\hat{f},\hat{f}_B\}}\sup_{\bx_1,\bx_2} 
        [1-\{f(\bx_1)-f(\bx_2)\}]_+ \le 1 + 2 \sup_{f \in \{\hat{f},\hat{f}_B\}}\sup_{1 \le i \le n}|f(\bx_i)|.
\end{align*}
Since $\hat{f}, \hat{f}_B \in \mathcal{H}$, the reproducing property and the Cauchy--Schwarz inequality yield
\begin{align*}
    \sup_{f \in \{\hat{f},\hat{f}_B\}}\sup_{1 \le i \le n}|f(\bx_i)| = \sup_{f \in \{\hat{f},\hat{f}_B\}}\sup_{1 \le i \le n}|\langle f, K(\cdot,\bx_i)\rangle_\mathcal{H}| \le \sup_{f \in \{\hat{f},\hat{f}_B\}}\|f\|_{\mathcal{H}}\sup_{1 \le i \le n}\|K(\cdot,\bx_i)\|_{\mathcal{H}} \le K_{\max}\sqrt{C_{\max}},
\end{align*}
which gives
\[
0 \;\le\; \sum_{I \in \Lambda} \zeta_k(I)H_f(\bx_{I_1},\bx_{I_2}) 
\;\le\;  1+2K_{\max}\sqrt{C_{\max}}.
\]
Therefore, 
Hoeffding's inequality yields
\begin{align*}
    P\left(\left|\frac{1}{B}\sum_{k=1}^{B}\sum_{I \in \Lambda} \zeta_k(I)H_f(\bx_{I_1},\bx_{I_2})
- \frac{1}{|\Lambda|}\sum_{I \in \Lambda}H_f(\bx_{I_1},\bx_{I_2})\right| > \eta \,\Big|\,(\bx_{I_1},\bx_{I_2})_{(I_1,I_2) \in \Lambda}\right)  \le 2\exp\!\left(-\frac{2B\eta^2}{(1+2K_{\max}\sqrt{C_{\max}})^2}\right).
\end{align*}
Consequently,
\[
P\left(\sup_{f \in \{\hat{f},\hat{f}_B\}} |\tilde U_B(f)-U_n(f)| > \eta \,\Big|\,(\bx_{I_1},\bx_{I_2})_{(I_1,I_2) \in \Lambda}\right) 
\le 4\exp\!\left(-\frac{2B\eta^2}{(1+2K_{\max}\sqrt{C_{\max}})^2}\right).
\]
That is, for all $\delta \in (0,1)$, with probability at least $1-\delta$,
\[
\sup_{f \in \{\hat{f},\hat{f}_B\}} |\tilde U_B(f)-U_n(f)| 
\le (1+2K_{\max}\sqrt{C_{\max}})\sqrt{\frac{\log(4/\delta)}{2B}}.
\]
Therefore, combining with \eqref{proof:first_term}, we obtain that for all $\delta \in (0,1)$, with probability at least $1-\delta$,
\[
|U_n(\hat{f})-\tilde U_B(\tilde{f}_B)| 
\le (1+2K_{\max}\sqrt{C_{\max}})\sqrt{\frac{\log(4/\delta)}{2B}}.
\]

For the second term $|\tilde U_B(\hat{f}_B)-\tilde U_B(\tilde{f}_B)|$, 
\begin{align*}
     |\tilde U_B(\hat{f}_B)-\tilde U_B(\tilde{f}_B)| &\quad\le \frac{1}{B} 
       \sum_{(i,j) \in \mathcal{D}_B}
       \left| 
            \big[ 1 - \{ \hat{f}_B(\bx_i) - \hat{f}_B(\bx_j) \} \big]_{+}-\big[ 1 - \{ \tilde{f}_B(\bx_i) - \tilde{f}_B(\bx_j) \} \big]_{+}
       \right| \\
   &\quad\le \frac{1}{B} 
       \sum_{(i,j) \in \mathcal{D}_B}
       \big| \hat{f}_B(\bx_i) - \hat{f}_B(\bx_j)
           - \big( \tilde{f}_B(\bx_i) - \tilde{f}_B(\bx_j) \big) 
       \big| \quad \text{(1-Lipschitz property of hinge loss)} \\
          &\quad\le \frac{1}{B} 
       \sum_{(i,j) \in \mathcal{D}_B} \left(|\hat{f}_B(\bx_i)-\tilde{f}_B(\bx_i)|+|\hat{f}_B(\bx_j)-\tilde{f}_B(\bx_j)|\right)
\end{align*}
Here, we introduce a random variable $N_i \ (1 \le i \le n)$, which represents the number of times the $i$th sample is selected in $\mathcal{D}_B$ and let $N_{\max} = \max_{1 \le i \le n} N_i$.
Then, 
\begin{align*}
   \frac{1}{B} 
       \sum_{(i,j) \in \mathcal{D}_B} \left(|\hat{f}_B(\bx_i)-\tilde{f}_B(\bx_i)|+|\hat{f}_B(\bx_j)-\tilde{f}_B(\bx_j)|\right) & = \frac{1}{B}\sum_{1\le i \le n} N_i |\hat{f}_B(\bx_i)-\tilde{f}_B(\bx_i)| \\
   &\quad \le \frac{N_{\max}}{B}\sum_{1\le i \le n} |\hat{f}_B(\bx_i)-\tilde{f}_B(\bx_i)| \\
      &\quad \le \frac{N_{\max}}{B}\sum_{1\le i \le n} \left(|(\bK\hat{\btheta}_B)_i-(\bK \tilde{\btheta}_B)_i|+|(\bK \tilde{\btheta}_B)_i-\tilde \bK\tilde{\btheta}_B)_i|\right)
\end{align*}
Define $\Delta\btheta=\hat{\btheta}_B-\tilde{\btheta}_B$ and $\mathcal{E}=\bK-\tilde{\bK}$. Then,
\begin{align*}
   \frac{1}{B} 
       \sum_{(i,j) \in \mathcal{D}_B} \left(|\hat{f}_B(\bx_i)-\tilde{f}_B(\bx_i)|+|\hat{f}_B(\bx_j)-\tilde{f}_B(\bx_j)|\right) & \le \frac{N_{\max}}{B}(\|\bK\Delta\btheta\|_1+\|\mathcal{E} \tilde{\btheta}_B\|_1) \\
      &\quad \le \frac{N_{\max}}{B}(n\|\bK\Delta\btheta\|_{\infty}+\sqrt{n}\|\mathcal{E} \tilde{\btheta}_B\|_2) \\
      &\quad \le \frac{N_{\max}}{B}(n\|\bK\Delta\btheta\|_{\infty}+\sqrt{n}\|\mathcal{E}\|_2 \|\tilde{\btheta}_B\|_2)
\end{align*}
where the last inequality follows from the Cauchy--Schwarz inequality.
In order to bound $\|\bK\Delta\btheta\|_{\infty}$, define 
\[
f_{\Delta}(\bx) = \sum_{k=1}^n \Delta\btheta_k\, K(\bx, \bx_k),
\]
so that $f_{\Delta} \in \mathcal{H}$ and $f_{\Delta}(\bx_i)=(\bK\Delta\btheta)_i$ for $1 \le i \le n$.
By the reproducing property and the Cauchy--Schwarz inequality,
\begin{align*}
\|\bK\Delta\btheta\|_{\infty}=\sup_{1 \le i \le n} |f_{\Delta}(\bx_i)|=\sup_{1 \le i \le n} | \langle f_{\Delta}, K(\cdot, \bx_i) \rangle_{\mathcal H} | \le \| f_{\Delta} \|_{\mathcal H} \sup_{1 \le i \le n} 
     \|K(\cdot, \bx_i)\|_{\mathcal H} =K_{\max} \, \|f_{\Delta}\|_{\mathcal H}.
\end{align*}
Thus, we have
\begin{align}
    \label{eq:second_bound}
|U_n(\hat{f})-\tilde U_B(\tilde{f}_B)| 
& \le \frac{N_{\max}}{B}(nK_{\max} \, \|f_{\Delta}\|_{\mathcal H}+\sqrt{n}\|\mathcal{E}\|_2 \|\hat{\btheta}_B\|_2).
\end{align}
To bound $\|f_{\Delta}\|_{\mathcal H}=\|\Delta{\btheta}\|_{\mathcal{H}}=\sqrt{\Delta{\btheta}^{\top}\bK\Delta{\btheta}}$, consider the problem \eqref{eq:kernel_ROC_SVM_incomp}. 
As the problem~\eqref{eq:ROCSVM} can be reformulated into problem~\eqref{eq:kernel-ROCSVM}, 
the objective function in~\eqref{eq:kernel_ROC_SVM_incomp} can likewise be expressed in terms of $\btheta$:
\[
  F_{B,\bk}(\btheta) = L_{B,\bk}(\btheta) + \frac{\lambda}{2} \|\btheta\|^2_{\mathcal{H}},
\]
where
\[
  L_{B,\bk}(\btheta) = \frac{1}{B} \sum_{(i,j) \in \mathcal{D}_B} 
  \big[ 1 - \btheta^{\top}(\bk_i - \bk_j) \big]_+ .
\]
The function $L_{B,\bk}(\btheta)$ is convex and the regularization term 
$\frac{\lambda}{2} \|\btheta\|^2_{\mathcal{H}}$ is $\lambda$-strongly convex, which implies $F_{B,\bk}(\btheta)$ is $\lambda$-strongly convex.  
Since $\hat{f}_B$ is the solution of \eqref{eq:kernel_ROC_SVM_incomp}, $\hat{\btheta}_B$ obtained from \eqref{def:hat_f_b} minimizes $F_{B,\bk}(\btheta)$. Therefore, 
\begin{align}
\label{ineq:strong-convex}
      \frac{\lambda}{2} \|\Delta\btheta\|_{\mathcal{H}}^2 
  \le F_{B,\bk}(\tilde{\btheta}_B) - F_{B,\bk}(\hat{\btheta}_B).
\end{align}
To proceed, we decompose the right-hand side as
\begin{align*}
F_{B,\bk}(\tilde{\btheta}_B) - F_{B,\bk}(\hat{\btheta}_B) 
&= \big[ F_{B,\bk}(\tilde{\btheta}_B) - F_{B,\tilde{\bk}}(\tilde{\btheta}_B) \big] 
+ \big[ F_{B,\tilde{\bk}}(\tilde{\btheta}_B) - F_{B,\tilde{\bk}}(\hat{\btheta}_B) \big] + \big[ F_{B,\tilde{\bk}}(\hat{\btheta}_B) - F_{B,\bk}(\hat{\btheta}_B) \big],
\end{align*}
where $F_{B,\tilde{\bk}}(\btheta)$ and $L_{B,\tilde{\bk}}(\btheta)$ are defined analogously to 
$F_{B,\bk}(\btheta)$ and $L_{B,\bk}(\btheta)$, with $\bK,\bk$ replaced by their Nyström approximations $\tilde{\bK},\tilde{\bk}$. 
That is, $F_{B,\tilde{\bk}}(\btheta)$ is the objective function of \eqref{eq:kernel-ROCSVM-incomp} written in terms of $\boldsymbol{\theta}$, obtained by substituting 
$f(\bx) =\sum_{k=1}^n \theta_k \tilde K(\bx, \bx_k)$
into \eqref{eq:kernel-ROCSVM-incomp}.
Since $\tilde{f}_B$ is the solution of \eqref{eq:kernel-ROCSVM-incomp} and thus the corresponding $\tilde{\btheta}_B$ minimizes $F_{B,\tilde \bk}(\btheta)$, it follows that 
\[
  F_{B,\tilde{\bk}}(\tilde{\btheta}_B) - F_{B,\tilde{\bk}}(\hat{\btheta}_B) \le 0 .
\]
Therefore,
\begin{align*}
F_{B,\bk}(\tilde{\btheta}_B) - F_{B,\bk}(\hat{\btheta}_B)
&\le \big[ F_{B,\bk}(\tilde{\btheta}_B) - F_{B,\tilde{\bk}}(\tilde{\btheta}_B) \big] 
+ \big[ F_{B,\tilde{\bk}}(\hat{\btheta}_B) - F_{B,\bk}(\hat{\btheta}_B) \big] \\
&= \frac{1}{B}\sum_{(i,j) \in \mathcal{D}_B} 
\big\{\big[1 - \tilde{\btheta}^{ \top}_B(\bk_i - \bk_j )\big]_+ 
    - \big[1 - \tilde{\btheta}^{ \top}_B(\tilde{\bk}_i - \tilde{\bk}_j )\big]_+ \big\} \\
&\quad + \frac{1}{B}\sum_{(i,j) \in \mathcal{D}_B} 
\big\{\big[1 - \hat{\btheta}^{ \top}_B(\tilde{\bk}_i - \tilde{\bk}_j )\big]_+ 
    - \big[1 - \hat{\btheta}^{ \top}_B(\bk_i - \bk_j )\big]_+ \big\}.
\end{align*}

\noindent
Using the 1-Lipschitz property of the hinge loss,
\begin{align}
F_{B,\bk}(\tilde{\btheta}_B) - F_{B,\bk}(\hat{\btheta}_B)
&\quad \le \frac{1}{B} \sum_{(i,j)\in \mathcal{D}_B} \{|\tilde{\btheta}_B^{\top} (\bk_i-\bk_j)-\tilde{\btheta}_B^{\top} (\tilde \bk_i-\tilde \bk_j)|+|\hat{\btheta}_B^{\top} (\tilde \bk_i-\tilde \bk_j)-\hat{\btheta}_B^{\top} (\bk_i-\bk_j)|\}  \nonumber \\
&\quad = \frac{1}{B} \sum_{(i,j)\in \mathcal{D}_B} |(\bK\tilde{\btheta}_B)_i-(\bK\tilde{\btheta}_B)_j-(\tilde \bK\tilde{\btheta}_B)_i+(\tilde \bK\tilde{\btheta}_B)_j| \nonumber \\
& \qquad +\frac{1}{B}\sum_{(i,j)\in \mathcal{D}_B}|(\tilde \bK\hat{\btheta}_B)_i-(\tilde \bK\hat{\btheta}_B)_j-(\bK\hat{\btheta}_B)_i+(\bK\hat{\btheta}_B)_j| \nonumber \\
&\quad \le \frac{1}{B} \sum_{(i,j)\in \mathcal{D}_B} \{|(\bK\tilde{\btheta}_B)_i-(\tilde \bK\tilde{\btheta}_B)_i|+|(\bK\tilde{\btheta}_B)_j-(\tilde \bK\tilde{\btheta}_B)_j|\} \nonumber \\
& \qquad +\frac{1}{B}\sum_{(i,j)\in \mathcal{D}_B}\{|(\tilde \bK\hat{\btheta}_B)_i-(\bK\hat{\btheta}_B)_i|+|(\tilde \bK\hat{\btheta}_B)_j-(\bK\hat{\btheta}_B)_j|\} \nonumber \\
&\quad = \frac{1}{B} \sum_{l \le i \le n}N_i\{|(\bK\tilde{\btheta}_B)_i-(\tilde \bK\tilde{\btheta}_B)_i|+|(\tilde \bK\hat{\btheta}_B)_i-(\bK\hat{\btheta}_B)_i|\} \nonumber \\
&\quad \le \frac{N_{\max}}{B} \sum_{l \le i \le n}\{|(\bK\tilde{\btheta}_B)_i-(\tilde \bK\tilde{\btheta}_B)_i|+|(\tilde \bK\hat{\btheta}_B)_i-(\bK\hat{\btheta}_B)_i|\} \nonumber \\
&\quad \le \frac{N_{\max}}{B} \sum_{l \le i \le n}\{|(\mathcal{E}\tilde{\btheta}_B)_i|+|(\mathcal{E}\hat{\btheta}_B)_i|\} \nonumber \\
&\quad \le \frac{N_{\max}\sqrt{n}}{B} (\|\mathcal{E}\tilde{\btheta}_B\|_2+\|\mathcal{E}\hat{\btheta}_B\|_2) \nonumber \\
&\quad \le \frac{N_{\max}\|\mathcal{E}\|_2\sqrt{n}}{B} (\|\tilde{\btheta}_B\|_2+\|\hat{\btheta}_B\|_2)\label{ineq:delta_F}
\end{align}

\noindent
Applying \eqref{ineq:delta_F} into \eqref{ineq:strong-convex},
\[
  \frac{\lambda}{2} \|\Delta\btheta\|_{\mathcal{H}}^2 
  \le\frac{N_{\max}\|\mathcal{E}\|_2\sqrt{n}}{B} (\|\tilde{\btheta}_B\|_2+\|\hat{\btheta}_B\|_2),
  \]
and therefore
\begin{align}
\label{ineq:L2-bound}
      \|\Delta\btheta\|_{\mathcal{H}}
  \le \sqrt{\frac{2N_{\max}\|\mathcal{E}\|_2\sqrt{n}}{B\lambda} (\|\tilde{\btheta}_B\|_2+\|\hat{\btheta}_B\|_2)}.
\end{align}
Substituting \eqref{ineq:L2-bound} into \eqref{eq:second_bound},
\[|\tilde U_B(\hat{f}_B)-\tilde U_B(\tilde{f}_B)|
\le \frac{N_{\max}}{B}\left(\|\mathcal{E}\|_2 \|\tilde{\btheta}_B\|_2\sqrt{n}+K_{\max}\sqrt{\frac{2N_{\max}\|\mathcal{E}\|_2n^{3/2}}{B\lambda} (\|\tilde{\btheta}_B\|_2+\|\hat{\btheta}_B\|_2)}\right)
\]
Plugging in $\| \hat{\btheta} \|_2 \;\leq\; \sqrt{ \frac{ C }{ \lambda_{\min} } }$ and $\|\tilde{\btheta}_B\|_2 
\;\le\; \sqrt{\frac{\tilde{C}}{\tilde{\lambda}_{\min}^+}}$, we have
\[|\tilde U_B(\hat{f}_B)-\tilde U_B(\tilde{f}_B)|
\le \frac{N_{\max}}{B}\left(\|\mathcal{E}\|_2 \sqrt{\frac{n\tilde{C}}{\tilde{\lambda}_{\min}^+}}+K_{\max}\sqrt{\frac{2N_{\max}\|\mathcal{E}\|_2n^{3/2}}{B\lambda} \left(\sqrt{ \frac{ C }{ \lambda_{\min} } }+\sqrt{\frac{\tilde{C}}{\tilde{\lambda}_{\min}^+}}\right)}\right).
\]
In conclusion, for all $\delta \in (0,1)$, with probability at least $1 - \delta$, 
\begin{align*}
    |U_n(\hat{f})-\tilde U_B(\tilde{f}_B)| & \le (1+2K_{\max}\sqrt{C_{\max}})\sqrt{\frac{\log(4/\delta)}{2B}} \\  & + \frac{N_{\max}}{B}\left(\|\mathcal{E}\|_2 \sqrt{\frac{n\tilde{C}}{\tilde{\lambda}_{\min}^+}}+K_{\max}\sqrt{\frac{2N_{\max}\|\mathcal{E}\|_2n^{3/2}}{B\lambda} \left(\sqrt{ \frac{ C }{ \lambda_{\min} } }+\sqrt{\frac{\tilde{C}}{\tilde{\lambda}_{\min}^+}}\right)}\right).
\end{align*}
\end{proof}
\end{document}